\documentclass[10pt,times,mathptm,psfig,twocolumn,journals]{IEEEtran}
\usepackage{url}
\usepackage{amssymb}
\usepackage{array}
\usepackage{amsmath}
\usepackage{graphicx}
\usepackage{booktabs}
\usepackage{epsfig}
\usepackage{ulem}
\usepackage{hyperref}
\hypersetup{pdfborder={0 0 0}, colorlinks=true, linkcolor=red, citecolor=blue}
\usepackage{subfigure}
\usepackage{cite}
\usepackage{epstopdf}
\usepackage{amsmath}
\usepackage[misc]{ifsym}
\usepackage{bm}
\usepackage{mathrsfs}

\usepackage{setspace}
\usepackage{multirow}
\usepackage{tabularx}
\usepackage{url}
\usepackage{booktabs}

\begin{document}

\title{Learning Edge-Preserved Image Stitching from Large-Baseline Deep Homography}

\author{Lang~Nie, Chunyu~Lin,~\IEEEmembership{Member,~IEEE}, Kang~Liao,~\IEEEmembership{Student Member,~IEEE}, Yao~Zhao,~\IEEEmembership{Senior Member,~IEEE}
\thanks{This work was supported by the National Natural Science
Foundation of China (No.61772066, No.61972028).
\textit{(Corresponding author: Chunyu Lin)}}
\thanks{Lang Nie, Chunyu Lin, Kang Liao, Yao Zhao are with the Institute of Information Science, Beijing Jiaotong University, Beijing 100044, China, and also with the Beijing Key Laboratory of Advanced Information Science and Network Technology, Beijing 100044, China (email: nielang@bjtu.edu.cn, cylin@bjtu.edu.cn, kang\_liao@bjtu.edu.cn, yzhao@bjtu.edu.cn).}
}

\maketitle
\begin{abstract}
    Image stitching is a classical and crucial technique in computer vision, which aims to generate the image with a wide field of view. The traditional methods heavily depend on the feature detection and require that scene features be dense and evenly distributed in the image, leading to varying ghosting effects and poor robustness. Learning methods usually suffer from fixed view and input size limitations, showing a lack of generalization ability on other real datasets. In this paper, we propose an image stitching learning framework, which consists of a large-baseline deep homography module and an edge-preserved deformation module. First, we propose a large-baseline deep homography module to estimate the accurate projective transformation between the reference image and the target image in different scales of features. After that, an edge-preserved deformation module is designed to learn the deformation rules of image stitching from edge to content, eliminating the ghosting effects as much as possible. In particular, the proposed learning framework can stitch images of arbitrary views and input sizes, thus contribute to a supervised deep image stitching method with excellent generalization capability in other real images.
    Experimental results demonstrate that our homography module significantly outperforms the existing deep homography methods in the large baseline scenes. In image stitching, our method is superior to the existing learning method and shows competitive performance with state-of-the-art traditional methods.

\end{abstract}

\markboth{}
{Shell \MakeLowercase{\textit{et al.}}: Bare Demo of IEEEtran.cls for IEEE Transactions on Magnetics Journals}
\IEEEpeerreviewmaketitle


\begin{figure*}[!t]
   \centering
   \subfigure[The input of our deep image stitching framework: the reference image $I_A$ and the target image $I_B$.]
   {\includegraphics[width=0.42\textwidth]{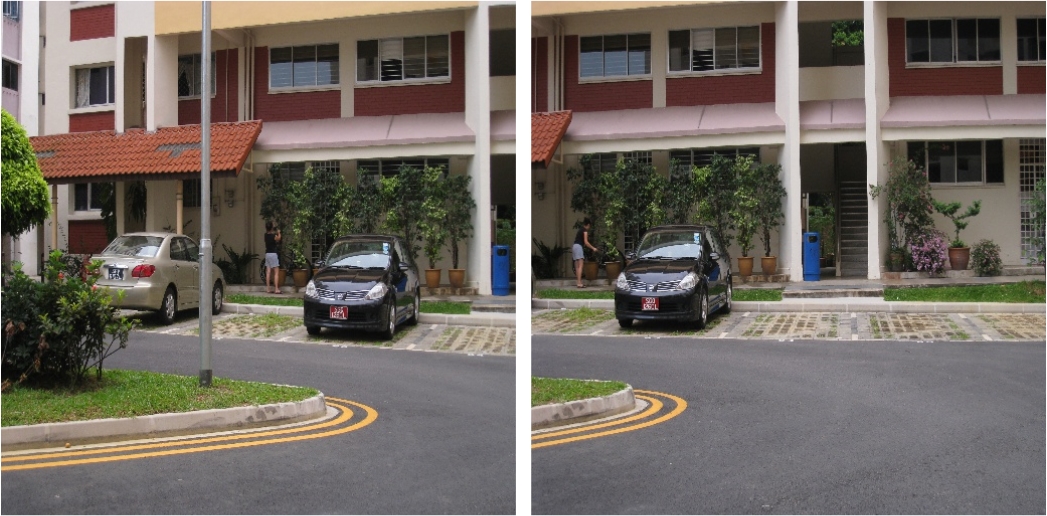}}
   \hspace{5ex}
   \subfigure[The output of proposed large-baseline deep homogrpahy module: warped reference image $I_{AW}$ and warped target image $I_{BW}$.]
   {\includegraphics[width=0.42\textwidth]{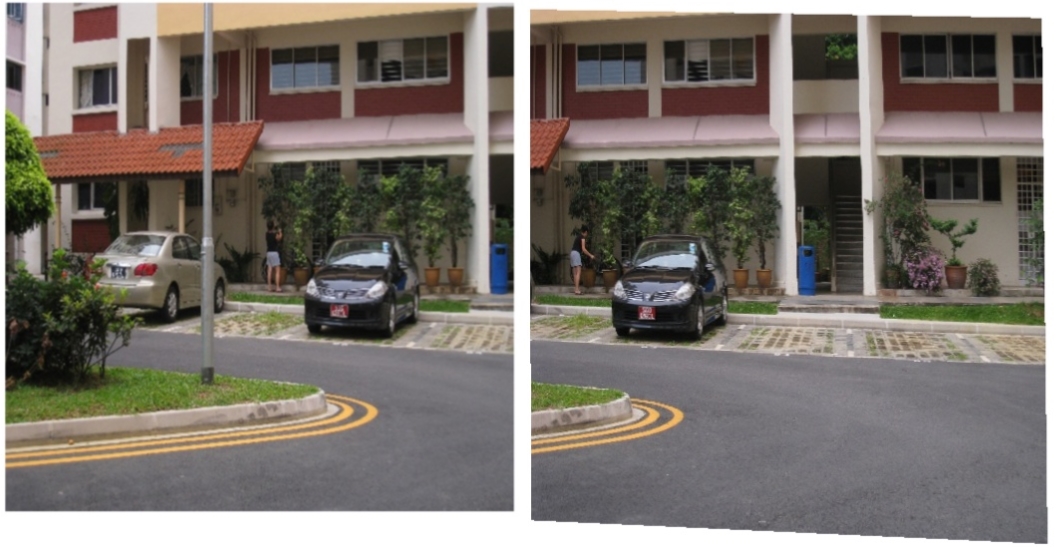}}
   \subfigure[Misalignments in overlapping areas caused by parallax.]
   {\includegraphics[width=0.31\textwidth]{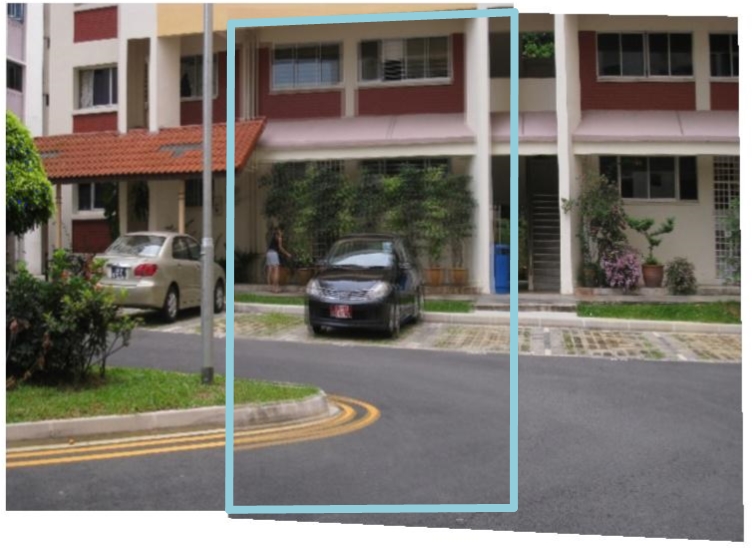}}
   \subfigure[Removing artifacts at the cost of edge discontinuity.]
   {\includegraphics[width=0.31\textwidth]{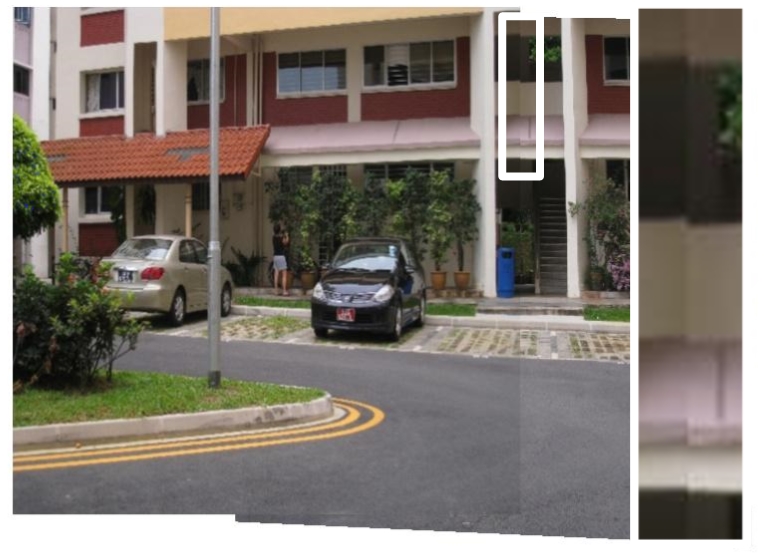}}
   \subfigure[The output of proposed edge-preserved deformation module: the stitched image with edge continuity correction.]
   {\includegraphics[width=0.31\textwidth]{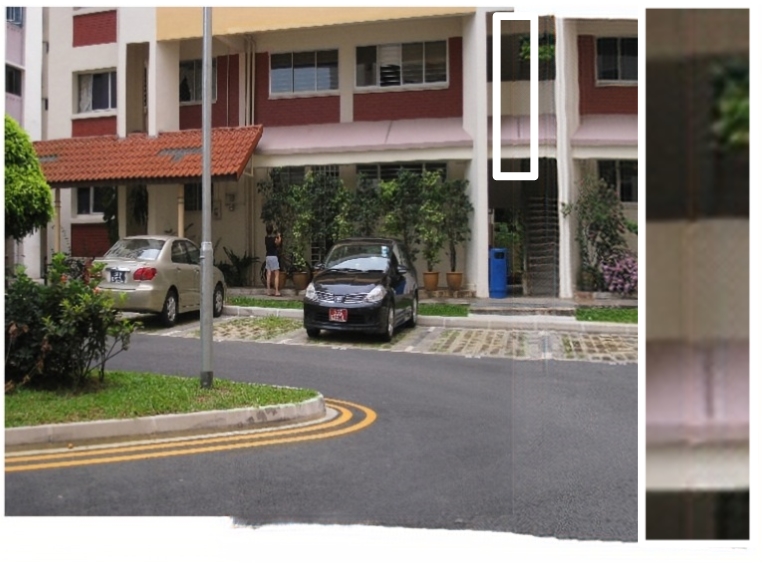}}
   \caption{The illustration of proposed edge-preserved image stitching strategy. 
   (a)(b) demonstrate the input and output of the large-baseline homography module, learning to align the large-baseline inputs coarsely.
   (c)(d)(e) exhibit the effect of the edge-preserved deformation module, learning to eliminate the artifacts and smooth the discontinuous edges simultaneously.}
   
   \label{LearningProcess}
\end{figure*}

\section{Introduction}
\label{section1}
\IEEEPARstart {D}UE to the limited field-of-view (FOV), a single photo cannot display the complete region of interest (ROI). To tackle this problem, a stitched image of a wider FOV can be obtained by stitching images from different viewing positions, which plays an important role in various applications such as autonomous driving \cite{wang2020multi, lai2019video}, immersive communication \cite{kasahara2016jackin}, virtual reality (VR) \cite{anderson2016jump, kim2019deep}.

Traditional image stitching methods follow similar steps: feature detection and matching, image registration, and image fusion. Among these steps, the most important is image registration that estimates a parametric transformation model from the target image domain into the reference image domain. Usually, the homography transformation is adopted, which can be effectively and simply represented as a $3\times3$ matrix. However, a single homography only contains the transformation from one plane to another \cite{hartley2003multiple} while the objects in an image are often at different depth levels. Hence, stitching with only a global homography 
frequently produces ghosting effects.

In order to mitigate ghosting effects, spatially-varying warping algorithms \cite{lou2014image, gao2011constructing, lin2011smoothly, zaragoza2013projective, chang2014shape, chang2012line, chen2016natural, lin2015adaptive, li2017parallax, lee2020warping, liu2018shape, li2019local, li2017quasi} have been proposed to learn spatially adaptive warpings. An image can be partitioned into different regions, and each region corresponds to a unique parametric transformation. By applying adaptive warpings to the target image, the overlapping areas of the images can be aligned to a considerable degree. Another category of traditional image stitching is the seam-driven image stitching \cite{eden2006seamless, zhang2014parallax, lin2016seagull, gao2013seam}. These methods search for the best seam-cut by minimizing the seam-related costs, reducing ghosting effects by seam-cut guided image fusion. However, these traditional methods' performance heavily depends on the condition that the feature points are dense and evenly distributed around the image, making these approaches not robust enough.

Recently, the deep learning methods have outperformed traditional methods in various computer vision tasks such as optical flow estimation \cite{sun2018pwc, dosovitskiy2015flownet, ilg2017flownet, truong2020glu, hu2018recurrent}, homography estimation \cite{detone2016deep, nguyen2018unsupervised, zhang2019content, le2020deep}. However, the deep image stitching is still in development. In deep image stitching, some methods are specially designed for fixed shooting positions \cite{lai2019video, shen2019real, li2019attentive} while some are implemented with convolutional neural networks (CNNs) applied in feature detection \cite{lai2019video, shen2019real, li2019attentive}, which cannot be regarded as complete deep image stitching algorithms. In addition, a view-free image stitching network (VFISNet) is proposed in \cite{nie2020view}, which successfully stitches images with arbitrary views in a complete deep learning framework for the first time. However, it has the limitations of fixed input size and weak generalization ability.

Considering those above traditional and learning methods' limitations, we propose a novel deep image stitching framework to stitch images from arbitrary views and input sizes in a flexible learning way. The proposed framework is composed of a large-baseline deep homography module and an edge-preserved deformation module. The first module achieves the homography estimation and image registration, and the remaining module learns the deformation rules of image stitching from edge to content.


For the crucial homography estimation stage, we found the following two common problems in the existing learning methods \cite{detone2016deep, nguyen2018unsupervised, zhang2019content, le2020deep}: 1) The learning process is only supervised at a single level. The above methods only use the features of the last convolution to predict the homography, while they ignore the different levels of features learned by other convolutional layers. As a result, the utilization of deep features is insufficient, and the network is hard to estimate an accurate projective transformation with the single-scale feature. 2) Learning the matching relationship of features by convolutional layers is inefficient, making these methods fail to work in large baseline scenes. In these methods, the receptive field of convolutional layer is limited by the kernel size, while the distance between matched features can be much larger than it.

To address the above problems, we propose a large-baseline deep homography module. In this module, we first adopt the feature pyramid to extract multi-scale features from coarse to fine. Then the feature correlation is implemented for the feature matching from global to local. Our network's receptive field can be significantly extended by combining feature pyramid with feature correlation, enabling our method to estimate the homography, especially in a large baseline. The input images (Fig. \ref{LearningProcess} (a)) can be warped using this estimated homography.


Subsequently, we design an edge-preserved deformation module to stitch the warped images (Fig. \ref{LearningProcess} (b)) from edge to content. Different from the traditional image stitching methods that aim to align the images as much as possible, our method learns the deformation rules of image stitching with the edge-preserved strategy. Because our framework is trained in a supervised manner using a no-parallax synthetic dataset, our framework learns to generate the overlapping areas of the stitched image only from the warped reference image, thus producing no artifact in the stitched image. However, as shown in Fig. \ref{LearningProcess} (d), learning the overlapping pixels only from the warped reference image would produce discontinuities in the edges between the warped reference image and the non-overlapping areas of the warped target image. Our edge-preserved deformation module overcomes this problem by learning to correct the discontinuity around the edges (Fig. \ref{LearningProcess} (e)), contributing to a visually pleasing and edge-continuity stitched result.

In experiments, we evaluate our method on the tasks of homography estimation and image stitching. Experimental results show that our approach outperforms previous methods with a large margin, demonstrating its robustness and efficacy on deep homography estimation and deep image stitching. The contributions of this paper are summarized as follows:

\begin{itemize}
   \item We design a large-baseline deep homography model, which adopts the feature pyramid and feature correlation simultaneously for the first time. Unlike the existing deep methods that estimate the homography in small-baseline scenes, the proposed approach is specially designed for large-baseline homography estimation, laying a solid foundation for deep image stitching.
   \item We propose an edge-preserved deformation network to stitch the warped images, eliminating the ghosting effects and keeping the edge continuity of the stitched image simultaneously.
   \item In the case that the fully connected layers are inevitable in the proposed deep image stitching framework, we designed a flexible mechanism combining image scaling and homography scaling to stitch images of arbitrary size.

\end{itemize}

The remainder of this paper is organized as follows: The related work is demonstrated in Section \ref{section2}. Our proposed large-baseline deep homography module and edge-preserved deformation module are discussed in Section \ref{section3}. The experiments and conclusions are presented in Section \ref{section4} and Section \ref{section5}, respectively.

\section{Related work}
\label{section2}
In this section, we review the traditional image stitching algorithms, deep homography estimation solutions, and deep image stitching methods.

\subsection{Traditional Image Stitching}
\label{section21}
\noindent\textbf{Spatially-Varying Warping.}
Traditional schemes stitch images with a single global homography, causing obvious ghosting effects \cite{hartley2003multiple}. To construct image panoramas with fewer artifacts, Gao $et\ al.$ proposed a dual-homography method (DHW) to represent the warpings of the foreground and background, respectively \cite{gao2011constructing}. To align different areas in the image domain, spatially adaptive warpings are calculated to stitch images as-projectively-as-possible (APAP) in the work of Zaragoza $et\ al.$ \cite{zaragoza2013projective}. Dividing pictures into dense grids, APAP calculates the spatially-adaptive warpings using moving DLT to seamlessly bridge image regions that are inconsistent with the projective model. However, the warping change of APAP in the adjacent areas is assumed to be small. In fact, the depth of the adjacent areas may change dramatically, which may still exhibit parallax artifacts in the vicinity of the object boundaries. Lee $et\ al.$ proposed the warping residual vectors to distinguish matching features from different depth planes \cite{lee2020warping}. More accurate stitching is achieved for images with large parallax by warping different patches with their corresponding estimated homography.

\begin{spacing}{1.5}
\end{spacing}

\noindent\textbf{Seam-Driven Methods.}
Seam-driven image stitching methods are also influential. A seam-cutting loss for the homography is proposed to measure the discontinuity between the warped target image and the reference image in the work of Gao $et\ al.$\cite{gao2013seam}. The homography with minimum seam-cutting loss is selected to achieve the best stitching. Zhang $et\ al.$ \cite{zhang2014parallax} introduced content-preserving warping (CPW) \cite{liu2009content} to align overlapping regions for small local adjustment while using the homography to maintain the global image structure. Different from aligning pixels of the overlapping area, Lin $et\ al.$ \cite{lin2016seagull} proposed to find a local area to stitch images, which can protect the curves and lines during stitching.

\subsection{Deep Homography Schemes}
\label{section22}
Homography estimation is an important part of image stitching, and deep homography can also be regarded as a significant step in deep image stitching. The deep homography solution was first proposed in \cite{detone2016deep} in 2016. In this work, a synthetic dataset for deep homography and a learning solution to predict the reference image's vertex displacements were put forward together. Then, Nguyen $et\ al.$ \cite{nguyen2018unsupervised} proposed an unsupervised solution for deep homography, in which a photometric loss is adopted to measure the pixel error between the warped target image and the reference image. In \cite{chang2017clkn}, a cascaded Lucas-Kanade Network is proposed to align images, where CNNs are used to extract multi-scale features and a Lucas-Kanade layer is utilized to find the motion parameters. Another multi-scale method is proposed in \cite{le2020deep}, which takes the image pyramid and self-attention mechanism into a learning framework at the same time. Zhang $et\ al.$ \cite{zhang2019content} propose a content-aware unsupervised solution, where a mask can be learned to work as an attention map to reject dynamic regions and select reliable areas for homography estimation simultaneously. This method achieved state-of-the-art performance for homography estimation in small baseline scenes.

\subsection{Deep Image Stitching}
\label{section23}
Deep image stitching is still in development, since the training dataset is hard to get, and the multi-task integrated stitching network is difficult to train. To reduce the learning burden of the network, some methods \cite{lai2019video, shen2019real, li2019attentive} designed a specific stitching situation such as a fixed camera shooting position, which cannot be extended to blind image stitching. Other methods \cite{hoang2020deep, shi2020image} adopted deep learning in a certain step of image stitching such as feature detection, which cannot be strictly regarded as a complete deep image stitching solution.
Besides that, the VFISNet was proposed in \cite{nie2020view}, where a cascaded network completely implemented by deep learning stitched images from arbitrary views. Nevertheless, this view-free network trained on synthetic dataset lacks the generalization ability, so it was difficult to stitch real images with parallax.

\section{Our method}
\label{section3}
In this section, we describe our proposed method in detail. First, we design a large-baseline deep homography module to achieve homography estimation and image registration in Section \ref{section31}. Then, an edge-preserved deformation network to stitch images with edge-preserved correction is proposed in Section \ref{section32}. Finally, some schemes to free the limitation of image size in deep image stitching are discussed in Section \ref{section33}.

\begin{figure*}[h]
   \centering
   \includegraphics[width=1\textwidth]{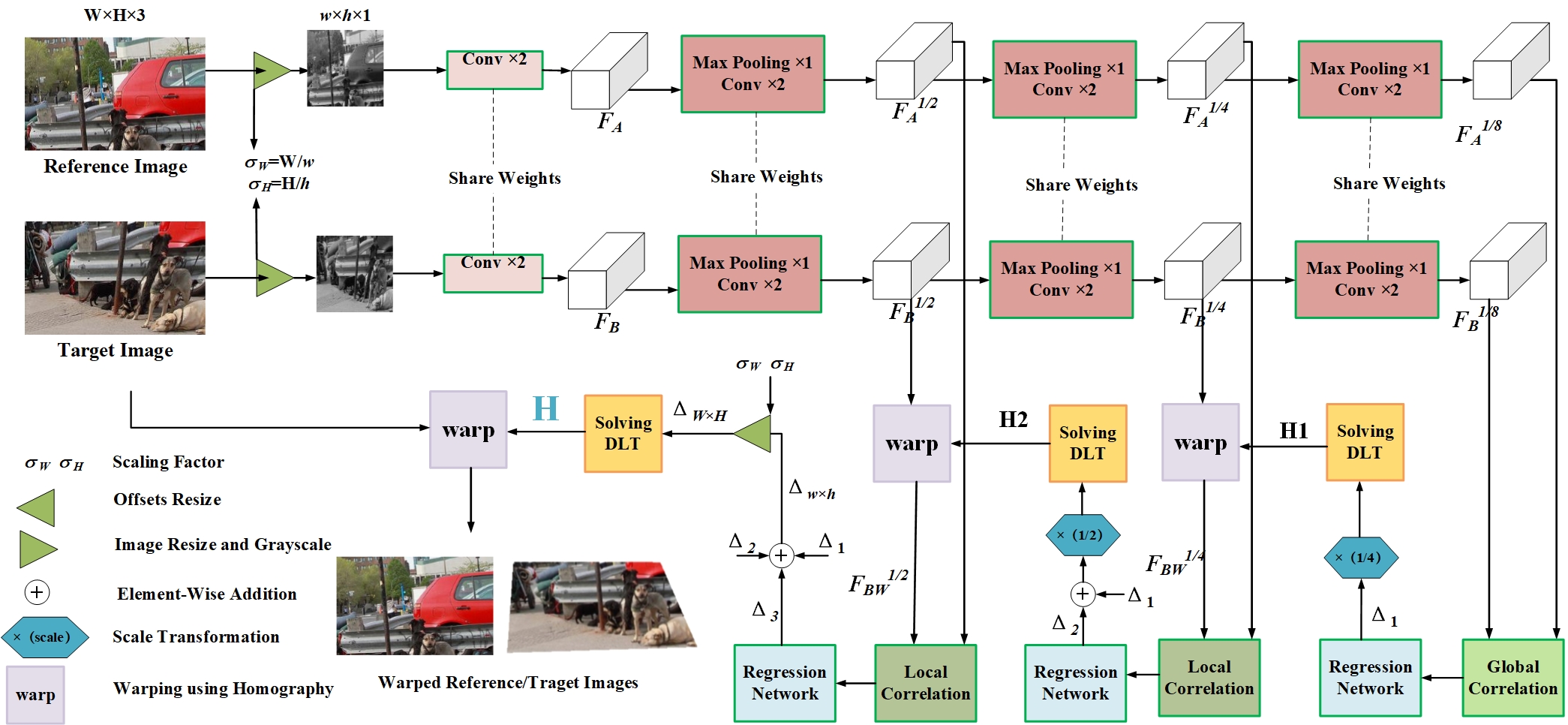}
   \caption{The architecture of our large-baseline deep homography network.}
   \label{MS-Homography}
\end{figure*}

\subsection{Large-Baseline Deep Homography}
\label{section31}

Although small-baseline deep homography methods \cite{detone2016deep, nguyen2018unsupervised, zhang2019content, le2020deep, chang2017clkn} have outperformed traditional homography solutions, large-baseline deep homography estimation is still challenging. Because in the scenes of a large baseline, the overlap rate between images is too low, and the receptive field of CNNs is significantly limited. To overcome this challenge, we propose a large-baseline deep homography network. In the field of deep homography, we combine feature pyramid and feature correlation into a network, increasing the utilization of feature maps and expanding our model's receptive field simultaneously.
In this manner, our network can perceive the correlation information in a large baseline, and the detailed architecture of our network is shown in Fig. \ref{MS-Homography}.

\begin{spacing}{1.5}
\end{spacing}

\noindent\textbf{Feature Pyramid.}
After the images are fed into our network, they will be processed by 8 convolutional layers, where the number of filters per layer is set to 64, 64, 128, 128, 256, 256, 512, and 512, respectively. A max-pooling layer is adopted every two convolutional layers to represent multi-scale features as $F$, $F^{1/2}$, $F^{1/4}$, and $F^{1/8}$. As shown in Fig. \ref{MS-Homography}, we select $F^{1/2}$, $F^{1/4}$, and $F^{1/8}$ to form a three-layer feature pyramid. The features of each layer in the pyramid are used to estimate the homography, and we transmit the estimated homography of the upper layer to the lower layer to continuously enhance the accuracy of the estimation. In this way, we predict the homography from coarse to fine in feature level.

\begin{spacing}{1.5}
\end{spacing}

\noindent\textbf{Feature Correlation.}
To increase the accuracy of homography estimation in the case of a large baseline, the feature correlation layer is used here to strengthen feature matching explicitly. Formally, the correlation $c$ between the reference feature $F_A^l\in W^l\times H^l\times C^l$ and the target feature $F_B^l\in W^l\times H^l\times C^l$ can be calculated as,
\begin{equation}
   c (x_A^l,x_B^l)=\frac{<F_A^l(x_A^l), F_B^l(x_B^l)>}{|F_A^l(x_A^l)||F_B^l(x_B^l)|},\ \ x_A^l,x_B^l\in \mathbb{Z}^2,
\end{equation}
where $x_A^l,x_B^l$ are the 2-D spatial location in $F_A^l$ and $F_B^l$, respectively. Specifying the search radius as $R$, we obtain $c\in W^l\times H^l\times (2R+1)^2$ by Eq. 1. Specifically, we calculate the global correlation by setting $R$ equal to $W^l$, and we calculate the local correlation when $R$ is less than $W^l$  (supposed that $W^l=H^l$). By applying global correlation and local correlation to our network, we predict the homography from global to local.

\begin{spacing}{1.5}
\end{spacing}

\begin{figure*}[t]
   \centering
   \includegraphics[width=0.97\textwidth]{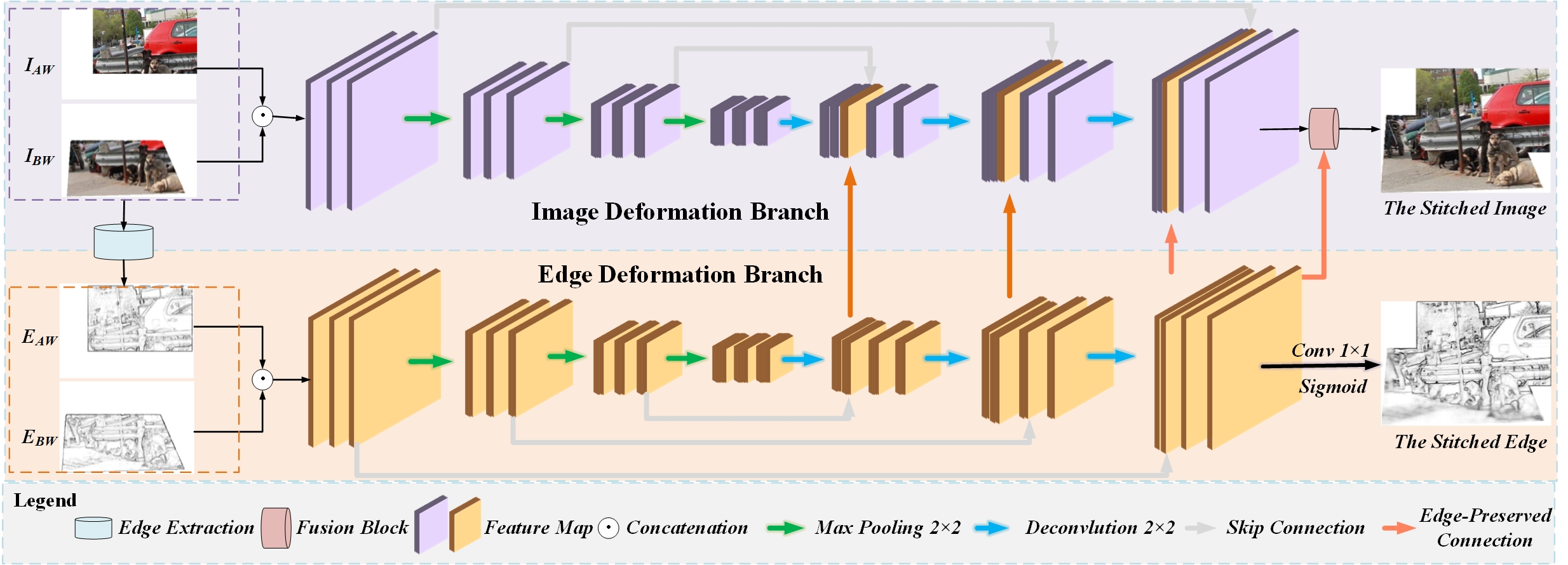}
   \caption{The architecture of edge-preserved deformation network. Top: Image deformation branch. Middle: Edge deformation branch. Bottom: Legend.}
   \label{EP-Stitching}
\end{figure*}

\begin{figure}[!t]
   \centering
      \includegraphics[width=0.5\textwidth]{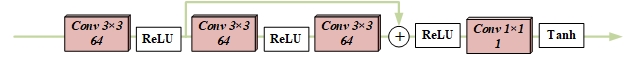}
      \caption{The detail of the fusion block in the edge-preserved stitching branch. }
      \label{Fusion-Block}
\end{figure}

After extracting pyramid features and calculating feature correlations, we adopt a simple regression network that comprises three convolutional layers and two fully connected layers to predict eight coordinate offsets that can uniquely determine a homography. To be more specific, every layer of our three-layer pyramid predicts the residual offsets $\Delta_i$, $i=1,2,3$.
Every feature correlation in the pyramid is only calculated between the warped target feature and the reference feature rather than between the target feature and the reference feature. In this way, each layer in the pyramid only learns to predict the residual homography offsets instead of the complete offsets.
And $\Delta_i$ can be calculated as follows:
\begin{equation}
   \Delta_{i} = \mathcal{H}_{4pt}\left\{ F_A^{1/2^{4-i}}, \mathcal{W} \left \langle F_B^{1/2^{4-i}}, \mathcal{DLT}(\sum_{n=0}^{i-1} \Delta_{n}) \right \rangle \right\},
\end{equation}
where $\mathcal{H}_{4pt}$ is the operation of estimating the residual offsets from the reference feature map and the warped target feature map. $\mathcal{W}$ warps the target feature map using the homography and $\mathcal{DLT}$ converts the offsets to the corresponding homography. We specify $\Delta_0=0$, which means all predicted offsets are 0. The final predicted offsets can be calculated as follows:
\begin{equation}
   \Delta_{w\times h} = \Delta_1 + \Delta_2 + \Delta_3.
\end{equation}
After that, image registration can be implemented by solving the homography and warping the input images.

\begin{spacing}{1.5}
\end{spacing}

\noindent\textbf{Objective Function:}
Our large-baseline deep homography is trained in a supervised manner. Given the ground truth offsets $\hat{\Delta_{w\times h}}$, we designed the following objective function,
\begin{equation}
    \begin{aligned}
   \mathcal{L}_{H} &= w_1(\hat{\Delta_{w\times h}}-\Delta_1) \\
   & + w_2(\hat{\Delta_{w\times h}}-\Delta_1-\Delta_2) \\
   & + w_3(\hat{\Delta_{w\times h}}-\Delta_1-\Delta_2-\Delta_3),
   \end{aligned}
\end{equation}
where the $w_1$, $w_2$, and $w_3$ represent the weights of each layer in the three-layer pyramid.

\subsection{Edge-Preserved Deformation Network}
\label{section32}

Stitching images with a global homography can easily produce artifacts. To eliminate the ghosting effects, we design an edge-preserved deformation network to learn the deformation rules of image stitching from edge to content. The learning process is quite different from traditional methods. As illustrated in Fig. \ref{LearningProcess} (d)(e), this learning method first eliminates all the artifacts at the cost of edge discontinuity and then learns to correct the discontinuity at the strategy of edge-preserved.


\begin{spacing}{1.5}
\end{spacing}

\noindent\textbf{Edge Deformation Branch.}
Compared with the rich information in an RGB image, such as color, texture, and content, the edge only contains the objects' contours in the image. Therefore, stitching the edges may be easier to achieve than stitching the RGB image. Inspired by this fact, we design an efficient and effective approach to extract edges, and an edge deformation branch is used to stitch them. The edge map $E$ for a grayscale image $G$ can be obtained by calculating the difference of adjacent pixels as follows,
\begin{equation}
   E_{i,j} = |G_{i,j}-G_{i-1,j}|+|G_{i,j}-G_{i,j-1}|,
\end{equation}
where $i$ and $j$ are the horizontal and vertical coordinates. A convolutional layer with fixed kernels can achieve the operation to extract edges. Finally, we clip $E_{i,j}$ between 0 and 1. As for the edge deformation branch, we implement it using an encoder-decoder architecture as shown in Fig. \ref{EP-Stitching} (middle). In this branch, the max pooling or deconvolution is adopted every two convolutional layers and the number of convolutional kernels is set to 64, 64, 128, 128, 256, 256, 512, 512, 256, 256, 128, 128, 64, 64, and 1, respectively. Among these convolutional layers, the size of all kernels is set to $3\times 3$ and the activation function is set to ReLU, except for the last convolutional layer. In the last layer, we set the kernel size to $1\times 1$ and the activation function as Sigmoid to generate the stitched edge. Furthermore, to prevent the gradient vanishing problem and information imbalance in the training \cite{ronneberger2015u}, skip connections are adopted to connect the low-level and high-level features with the same resolution.

\begin{spacing}{1.5}
\end{spacing}

\noindent\textbf{Image Deformation Branch.}
We also design an image deformation branch to generate the stitched image in the guidance of the stitched edges. The image deformation branch has a similar architecture to the edge deformation branch as shown in Fig. \ref{EP-Stitching} (top). To enable the image deformation branch of the edge-preserved stitching, we use the edge features learned by the edge deformation branch in the decoder stage to guide the learning.
To be specific, we concatenate each feature map obtained by deconvolution in the edge deformation branch with the corresponding feature map in the image deformation branch from low-level to high-level. Besides, a fusion block is designed to integrate the last feature map in the edge deformation branch with the corresponding feature map in the image deformation branch, as illustrated in Fig. \ref{Fusion-Block}.


\begin{spacing}{1.5}
\end{spacing}

\noindent\textbf{Objective Function.}
Similar to our deep homography, we train our stitching network in a supervised manner. To make the stitched edge close to the ground truth edge $\hat{E}$ that is extracted from the ground truth image $\hat{I}$, $\mathcal{L}_1$ loss is adopted as follows:
\begin{equation}
   \mathcal{L}_{edge}= \frac{1}{W\times H\times 1} \left \| \hat{E}-E \right \|_1,
\end{equation}
where $W$ and $H$ define the width and height of the stitched edge.

Inspired by \cite{johnson2016perceptual}, we define a content loss to encourage our image deformation branch to generate perceptual naturally stitched images. Specifically, we use the 9-th convolutional layer in VGG-19 \cite{simonyan2014very} as the representation of the image content. Let $\varPhi_j$ denotes the j-th layer of VGG-19 and we define our content loss as follows:
\begin{equation}
   \mathcal{L}_{content} = \frac{1}{W_j\times H_j\times C_j} \left \| \varPhi_j (\hat{I})-\varPhi_j (I) \right \|_2^2,
\end{equation}
where $W_j$ ,$H_j$ and $C_j$ denote the width, height, and channel number of the feature map, respectively.


Considering the constraints on the edge and content, we finally conclude our objective function as follows:
\begin{equation}
   \mathcal{L}_{S} = \lambda_e \mathcal{L}_{edge} + \lambda_c \mathcal{L}_{content},
\end{equation}
where the $\lambda_e$ and $\lambda_c$ represent the balance factors of edge loss and content loss, respectively.

\begin{figure}[!t]
   \centering
      \includegraphics[width=0.5\textwidth]{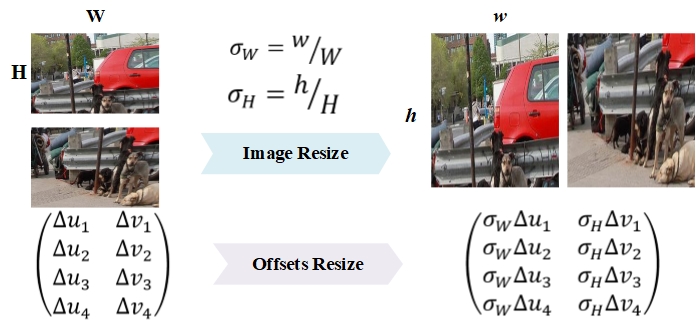}
      \caption{The relationship between image resize and offsets resize. $(\Delta u_i, \Delta v_i)$ represents the coordinate offsets of the i-th vertex in the target image, where i=1, 2, 3, and 4. }
      \label{Relationship}
\end{figure}

\subsection{Size-Free Stitching}
\label{section33}

Size-free image stitching can be easily achieved by replacing the fully connected layers with convolutional layers \cite{long2015fully}. However, the increase in input images' size will significantly increase the memory consumption because of feature correlation layers. Taking the global correlation as an example, the required memory can be expanded by $\lambda^4$ times when the size of input images is expanded by $\lambda$ times. To make it more clear, we show the change of memory consumption as follows,
\begin{equation}
   W^l\times H^l\times (2W^l+1)^2 \Rightarrow \lambda W^l\times \lambda H^l\times (2\lambda W^l+1)^2.
\end{equation}
Obviously, adopting a fully convolutional network (FCN) cannot solve this problem. To reduce endless memory consumption, we design an alternative to achieve size-free stitching.

When we resize the images, we can change the corresponding offsets following the rule shown in Fig. \ref{Relationship}. Noticing the relationship between image resize and offsets resize, we implement our size-free image stitching in three steps, as shown in Fig. \ref{MS-Homography}:
1)We resize the input images from $W\times H$ to $w\times h$ and save scaling factors for width and height $\sigma_W$, $\sigma_H$.
2)We predict the offsets from the images of $w\times h$.
3)We resize the offsets using $\sigma_W$ and $\sigma_H$ by the rule shown in Fig. \ref{Relationship} to make them correspond to the images of $W\times H$.
In short, we complete size-free homography estimation using the relationship between image resize and offsets resize without extra memory consumption. Since the edge-preserved deformation module can be regarded as an FCN, our deep image stitching framework can process arbitrary size inputs.

\section{Experiments}
\label{section4}
In this section, we carry out experiments to validate the effectiveness of our method. We first introduce our dataset and implementation details in Section \ref{section41}. Then, comparative experiments on homography estimation and image stitching are conducted in Section \ref{section42} and Section \ref{section43}, respectively. Finally, the ablation studies are formed in Section \ref{section44}.

\subsection{Dataset and Implementation Details}
\label{section41}

\begin{figure}[!t]
   \centering
      \includegraphics[width=8.5cm,height=1.95cm]{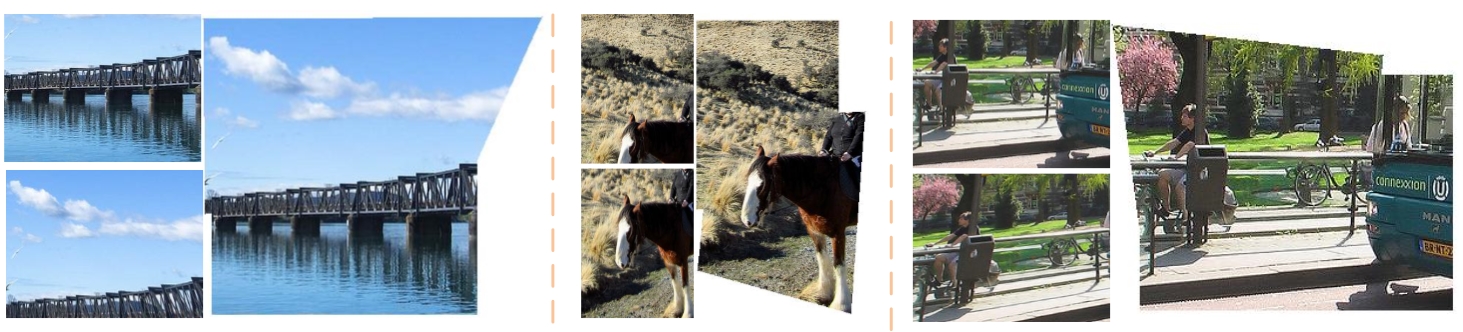}
      \caption{Several samples of our Stitched MS-COCO dataset. Each sample is separated by a dashed line. The $I_{Reference}$, $I_{Target}$ and $Label$ are demonstrated in each instance.}
      \label{dataset}
\end{figure}

\noindent\textbf{Dataset.}
Deep homography and deep image stitching are two different tasks, but we adopt the same dataset to train them together. We follow the strategy of \cite{nie2020view} to generate a seemingly infinite dataset for image stitching from Microsoft COCO \cite{lin2014microsoft}. We call this large-baseline dataset as Stitched MS-COCO, and we demonstrate some samples in Fig. \ref{dataset}. To be specific, in addition to the random perturbation $[-\rho, \rho]$ \cite{detone2016deep} of the four vertices in an image patch, the random translation $[-\tau, \tau]$ \cite{nie2020view} is added to simulate the characteristics of large baseline and low overlap in image stitching. The format of Stitched MS-COCO can be described as a quadruple $(I_{Reference}, I_{Target}, \Delta, Label)$, of which $I_{Reference}$ and $I_{Target}$ represent the reference image and target image to be stitched, $\Delta$ represents the 8 coordinate offsets of the four vertices to estimate a homography, and $Label$ is the ground truth of the stitched result. Specifically, when generating a quadruple from a real image ($W\times H$), we set the size of image patches ($P^W\times P^H$) to be input into our network to $W/2.4\times H/2.4$, the maximum translation ($\tau^W\times \tau^H$) to $0.5P^W\times 0.5P^H$, and the maximum perturbance ($\rho^W\times \rho^H$) to $0.2P^W\times 0.2P^H$. Moreover, $\Delta$ can be calculated by adding translation and perturbance. We generate 50,000 quadruples from MS-COCO train2014 as the training set and 5,000 quadruples from test2014 as the test set.

\begin{spacing}{1.5}
\end{spacing}

\noindent\textbf{Details.}
The training process is completed in two steps: deep homography module and deep deformation module successively. Our deep homography network is trained by an Adam optimizer \cite{kingma2014adam} for up to 100 epochs, with an exponentially decaying learning rate initialized as $10^{-4}$, a decay step of $12,500$, and a decay rate of $0.95$. According to the different influence of each pyramid layer on the homography prediction, we set $w_1$, $w_2$, and $w_3$ to 1, 0.25, and 0.1, respectively. We adopt some data augmentation techniques to enhance illumination robustness, such as artificially inserting random brightness shifts into the training images. Subsequently, we train our stitching module with the parameters of the homography network being fixed. The training strategy is the same as that of the homography module, except for the maximum training epoch being set as 25. The balance factors $\lambda_e$ and $\lambda_c$ are set to $1$ and $2e^{-6}$. In addition, the batch size numbers of the two training steps are set to 4 and 1. The input size $W\times H$ of our framework is arbitrary, and the scaling size $w\times h$ is set to $128\times 128$ which is consistent with \cite{detone2016deep, nguyen2018unsupervised, zhang2019content, le2020deep}. All the components of this framework are implemented on TensorFlow, and the training process is performed on one NVIDIA RTX 2080 Ti.

\subsection{Comparison with Homography Estimations}
\label{section42}
Traditional homography estimations differ according to different feature descriptors and different outlier rejections. Feature descriptor can be SIFT \cite{lowe2004distinctive}, ORB \cite{rublee2011orb}, and so on. The outlier rejection algorithm can be RANSAC \cite{fischler1981random}, MAGSAC \cite{barath2019magsac}, and so so. Since the combination of SIFT and RANSAC can reach better accuracy than other combinations \cite{nguyen2018unsupervised, zhang2019content}, we choose this combination as a representative of traditional solutions to compare. Besides that, we compare our method with deep homography algorithms, including DHN \cite{detone2016deep}, UDHN \cite{nguyen2018unsupervised}, and CA-UDHN \cite{zhang2019content}. When comparing the estimated homography with the ground truth, we adopt the same evaluation metric in \cite{nguyen2018unsupervised}, the 4pt-Homography RMSE.

\begin{table}
   \centering
   \caption{Comparison experiment for homography estimation on Warped MS-COCO $(\rho=32)$. The number represents the 4pt-Homography RMSE between the estimated offsets of 4 vertexs and the ground truth. All the learning methods are trained on Warped MS-COCO. $F$ indicates that this method fails in the current dataset.}
   \scalebox{0.73}{
      \begin{tabular}{llllll}
         \toprule
         Method & Top 0$\sim$30\% & 30$\sim$60\% & 60$\sim$100\% & Average\\
         \midrule
         $I_{3\times 3}$  & 15.0154 & 18.2515 & 21.3517 & 18.5220\\
         \midrule
         SIFT\cite{lowe2004distinctive}+RANSAC\cite{fischler1981random} & 0.6743 & 1.0964 & 19.0286 & 9.4782\\
         \midrule
         DHN\cite{detone2016deep} & 3.2998 & 4.8839 & 7.6944 & 5.5358\\
         \midrule
         UDHN\cite{nguyen2018unsupervised} & 2.1894 & 3.5272 & 6.4984 & 4.3179\\
         \midrule
         CA-UDHN\cite{zhang2019content} & $F$ & $F$ & $F$ & $F$\\
         \midrule
         Ours & $\mathbf{0.2719}$ & $\mathbf{0.4140}$ & $\mathbf{0.9632}$ & $\mathbf{0.5962}$\\
         \bottomrule
         \hline
         \end{tabular}
   }
         \label{HomographyCompare_1}
\end{table}

\begin{figure}[!t]
   \centering
      \includegraphics[width=0.5\textwidth]{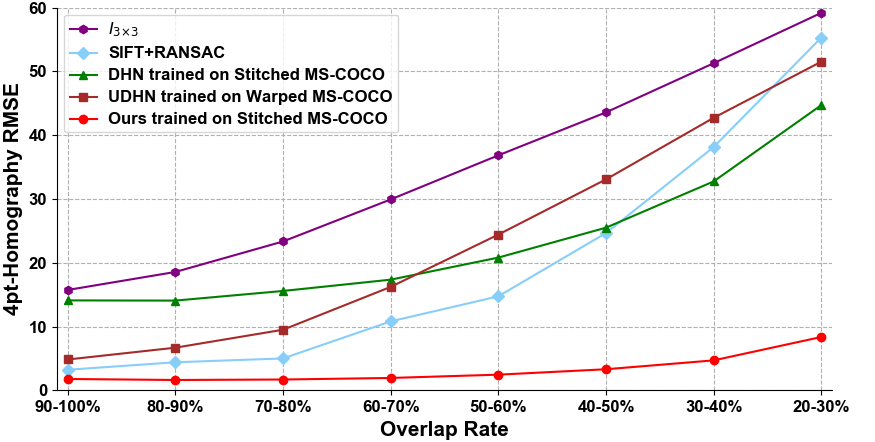}
      \caption{Comparative experiment for homography estimation on Stitched MS-COCO $(\tau=64, \rho=25)$.} 
      \label{HomographyCompare_2}
\end{figure}

\begin{spacing}{1.5}
\end{spacing}

\noindent\textbf{Warped MS-COCO.}
Warped MS-COCO, which only includes the random perturbance $[-\rho, \rho]$ of four vertices, is the most widely acknowledged synthetic dataset for deep homography estimation. We first conduct a comparative experiment on this dataset with $\rho=32$, where each corner of the image patch can be perturbed by a maximum of one-quarter of the total image size. The results are shown in Table \ref{HomographyCompare_1}, where $I_{3\times 3}$ refers to a ${3\times 3}$ identity matrix as a 'no-warping' homography for reference. The performance of traditional homography solution heavily relies on the quality of feature matching, which indicates this method may fail when the number of matched features is small or the matching accuracy is low. To avoid this problem, we set the estimated homography to the identity matrix when that happens. As shown in Table \ref{HomographyCompare_1}, the results are divided into several parts to illustrate each method's various performance profiles. Specifically, the method of SIFT and RANSAC performs pretty well in 60\% of all the test sets, while it usually cannot capture enough matching features to estimate a homography in the worst 40\% of all. UDHN and DHN achieve similar performance with offsets' error are controlled to several pixels all the time. CA-UDHN achieves state-of-the-art performance in small baseline scenes, while its performance is close to $I_{3\times 3}$ in large baseline scenes. This method fails to work because its perception field is limited, making it unable to perceive the two images' alignment information. Our large-baseline deep homography solution outperforms all the compared deep solutions and traditional methods with a large margin all the time.


\begin{spacing}{1.5}
\end{spacing}

\begin{figure*}[!t]
   \centering
   \includegraphics[width=1\textwidth]{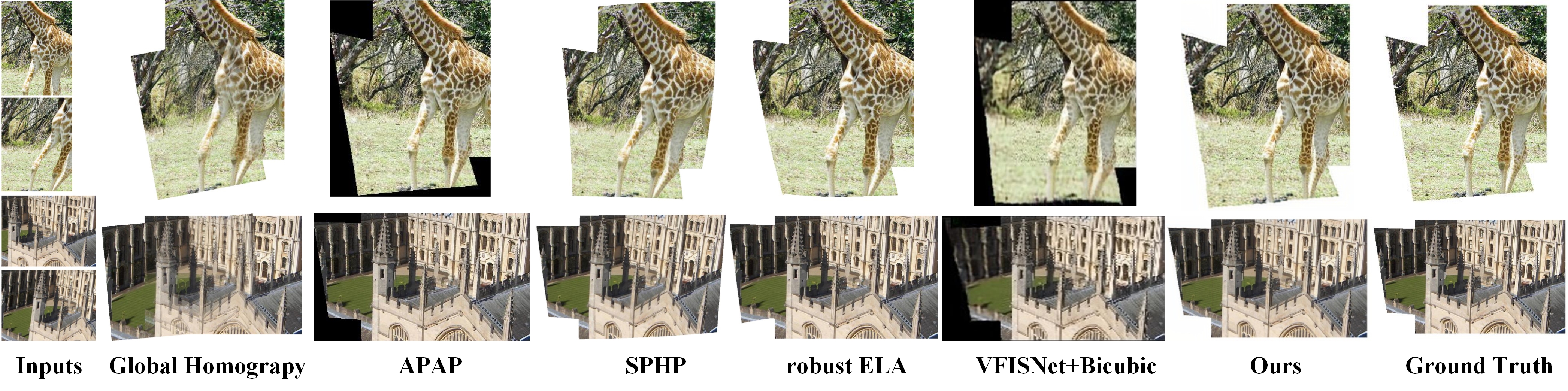}
   \caption{The comparative experiments in our synthetic dataset. Col 1: Input images. Col 2-7: Stitched results of the global homography, SPHP \cite{chang2014shape}, APAP \cite{zaragoza2013projective}, robust ELA \cite{li2017parallax}, VFISNet\cite{nie2020view}+Bicubic, and ours. Col 8: The ground truth.}
   \label{systhetic}
\end{figure*}

\begin{figure}[!t]
   \centering
   \includegraphics[width=0.45\textwidth]{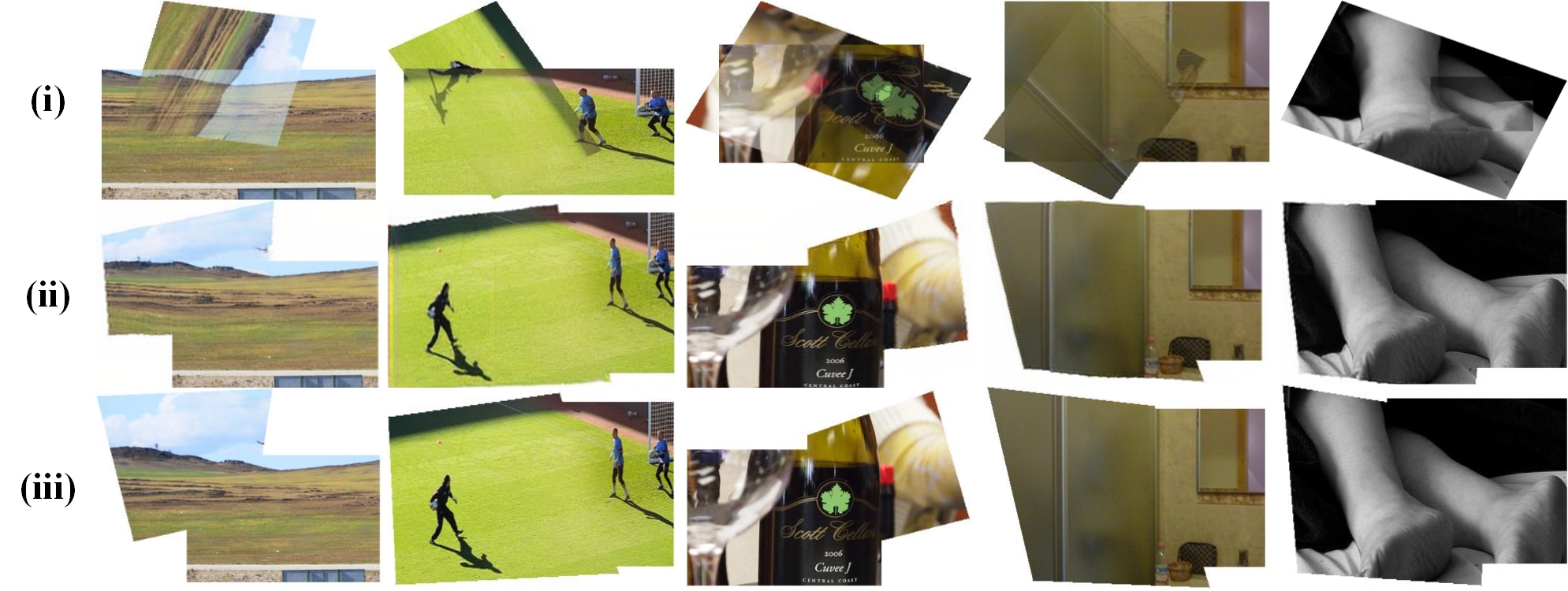}
      \caption{Failure cases of traditional feature-based methods. (i)(ii): The stitched results of the Global Homography and ours. (iii): The ground truth.}
      \label{systhetic-failure}
\end{figure}

\noindent\textbf{Stitched MS-COCO.}
In image stitching, the baseline between images is usually larger than that in Warped MS-COCO. Here, the existing homography estimation solutions' performance drops sharply as the baseline increases, while our method is still robust and accurate. We verified this view on Stitched MS-COCO dataset that is much more challenging due to the larger displacement and the lower overlap rate. To be consistent with Warped MS-COCO, we resize $I_{Reference}$ and $I_{Target}$ to $128\times 128$ in this experiment. Compared with the supervised solution DHN, the unsupervised solution UDHN requires extra information around the image patch to prevent ambiguity during the training process \cite{nguyen2018unsupervised, zhang2019content}. However, Stitched MS-COCO is only composed of image patches and corresponding homography offsets, which makes UDHN unable to be trained on this dataset. Therefore, we test UDHN using the model trained on Warped MS-COCO. The results are shown in Fig. \ref{HomographyCompare_2}. As the overlap rate decreases, the accuracy of all methods continues to decrease, of which the accuracy of SIFT+RANSAC, DHN, and UDHN decreases faster than our method significantly. And the lower the overlap rate is, the closer the performance of the three methods is to $I_{3\times 3}$, which indicates that these methods may fail to work when the overlap rate is particularly low. In contrast, our method can maintain good accuracy even at low overlap rates, which lays a solid foundation for image stitching.

\begin{spacing}{1.5}
\end{spacing}

From comparative experiments on Warped MS-COCO and Stitched MS-COCO, it can be observed that our large-baseline deep homography outperforms the existing deep solutions and traditional solutions, especially in large baseline scenes. By combining feature pyramid and feature correlation, the homography can be accurately estimated from coarse to fine and from global to local.

\subsection{Comparison with Image Stitching Algorithms}
\label{section43}
Most deep image stitching algorithms are specially designed for a specific task \cite{lai2019video, shen2019real, li2019attentive} or can not be regarded as a complete deep learning framework \cite{hoang2020deep, shi2020image}. Therefore, it is not fair or convincing to compare our algorithm with them. Instead, we choose VFISNet \cite{nie2020view}, a complete view-free image stitching network, as a representative of deep image stitching to compare. Since its input size is $128\times 128$, we combine it with Bicubic interpolation to produce the stitched results of arbitrary size. As for traditional methods, we compare our method with four classical image stitching algorithms: Global Homography, SPHP \cite{chang2014shape}, APAP\cite{zaragoza2013projective}, and robust ELA \cite{li2017parallax}, in which the first two are classic methods with global transformation models and the others are with local adaptive stitching fields. Among these four methods, we implement Global Homography using SIFT, RANSAC, and average fusion. The results of SPHP, APAP, and robust ELA are obtained by running their open-source codes with our testing instances. These methods are evaluated on our synthetic images and real images, respectively.


\begin{spacing}{1.5}
\end{spacing}



\noindent\textbf{Synthetic Images.}
The stitched results in our synthetic dataset are illustrated in \ref{systhetic}. There are obvious artifacts in the stitched result of Global Homography because the mismatch of feature points affects homography estimation accuracy. Compared with SPHP, APAP, and robust ELA, our solution shows competitive performance with these classic and convincing image stitching works. In deep image stitching methods, our results are more visually clear than that of VFISNet+Bicubic.

Besides that, our method is more robust. Traditional methods heavily depend on the quality of feature detection and feature matching. However, the feature points can be easily affected by various environments. We test 1,000 pairs of images in our test set with the Global Homography and our method. Experimental results show that more than 30 pairs fail using the Global Homography, while all work in our method. Fig. \ref{systhetic-failure} shows some failure cases of traditional methods in our synthetic dataset. As for other feature-based methods, the number of failures can be several times as that of the Global Homography, because they usually have stricter requirements on the distribution or number of feature points. For instance, APAP would require more feature points to find a valid point subset when generating hypotheses for multi-structure data \cite{chin2011accelerated}. The robustness of our method benefits from the powerful feature extraction capability of CNNs, which has been proven in other similar fields such as optical flow estimation \cite{sun2018pwc, dosovitskiy2015flownet, ilg2017flownet, truong2020glu}.

\begin{spacing}{1.5}
\end{spacing}

\begin{figure*}[!t]
   \centering
   \includegraphics[width=1\textwidth]{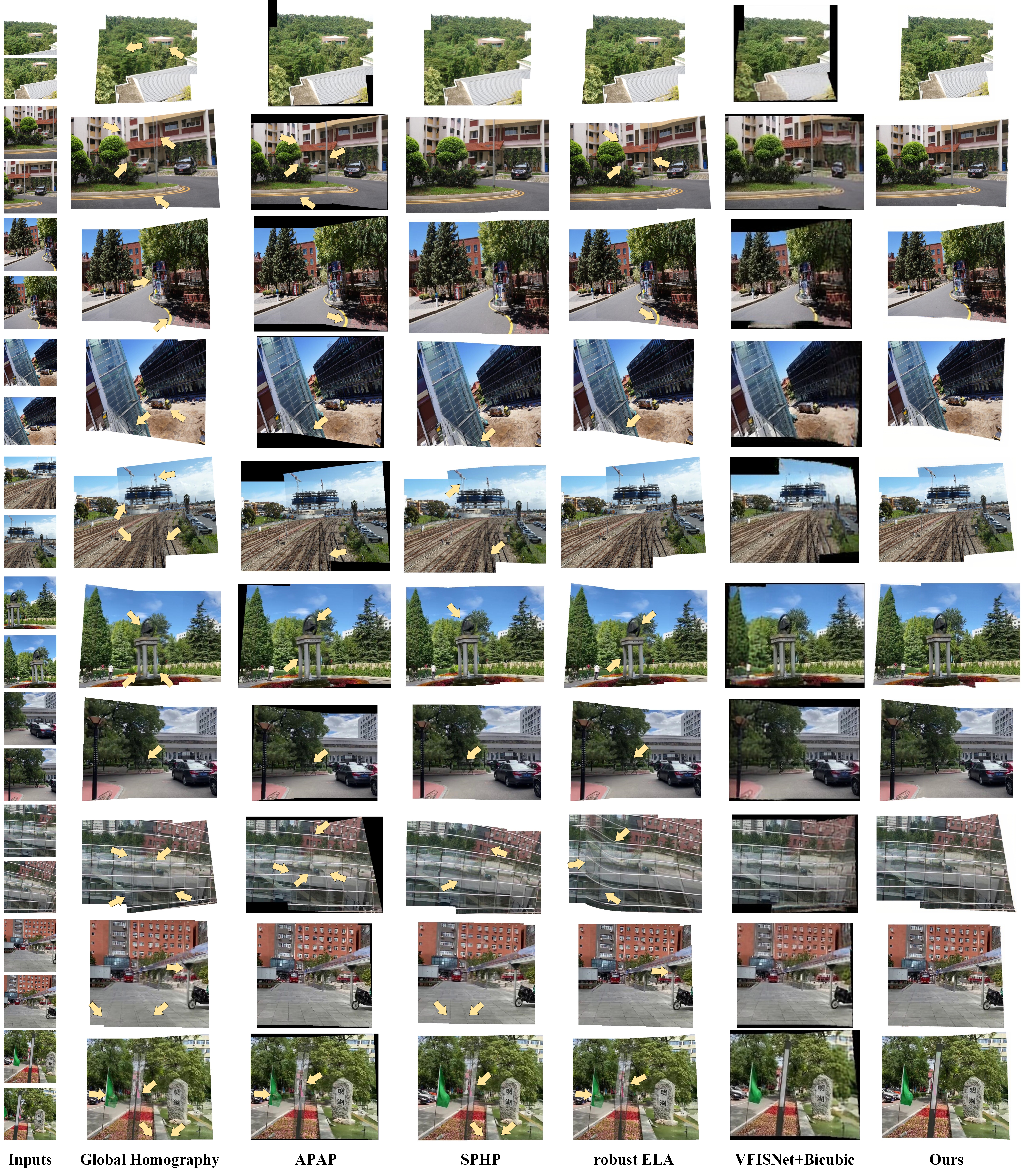}
   \caption{The comparative experiments in real images. Col 1: Input images. Col 2-7: Stitching results of the global homography, SPHP \cite{chang2014shape}, APAP \cite{zaragoza2013projective}, robust ELA \cite{li2017parallax}, VFISNet\cite{nie2020view}+Bicubic, and ours. The first 5 examples come from classic image stitching cases, and the last 5 are challenging cases with obvious parallax or even moving person taken by ourselves.}
   \label{natural}
\end{figure*}

\noindent\textbf{Real Images.}
In addition to synthetic images, we also test our model on real images with apparent parallax. Although our method is merely trained on a synthetic dataset without parallax, it can produce perceptually natural stitched results even in real images, contributing to a supervised deep image stitching method with excellent generalization capability. It benefits from our learning framework, where the first module learns to align the images coarsely and the second module learns to generate a stitched image with no artifact.

As shown in Fig. \ref{natural}, the first 5 examples come from classic image stitching cases that are widely used in existing traditional image stitching methods, and the last 5 are challenging cases with obvious parallax or even moving objects taken by ourselves. The arrows highlight the artifacts. Due to GPU memory limitation, we limit the input images' maximum size not to exceed $512\times 512$.
From the results shown in Fig. \ref{natural}, we can observe:

(1) The learning image stitching methods (VFISNet and ours) can eliminate almost all the artifacts, while the traditional methods (Global Homography, SPHP, APAP, robust ELA) cannot do it in various stitching scenes. This can be accounted for different stitching strategies. To eliminate the artifacts, the traditional solutions try to align the reference image and target image as much as possible.  However, the stitching quality heavily relies on the number and distribution of the feature points, failing to eliminate the ghosting effects in varying scenes. As for the proposed deep image stitching, the network tends to learn the overlapping areas from the reference image, neglecting the target image and free from the artifacts. Although this learning tendency may make the edges discontinuous, our network would learn to revise it to look smooth and natural.

(2) Our method outperforms the existing deep image stitching method. Although the deep solutions can eliminate the artifacts, they bring another problem: the stitched images' non-overlapping regions are blurred and discontinuous. This problem can be observed obviously in the results of VFISNet+Bicubic, while our method alleviates this problem by learning image stitching from edge to content progressively.

(3) In a scene containing moving objects, the learning methods perform better than the traditional methods. Row 7 of Fig. \ref{natural} exhibits a pair of images that contains a moving person. We can see that the Global Homography, SPHP, APAP, and robust ELA cannot handle this moving person while the learning methods deal with it successfully. 



\subsection{Ablation Studies}
\label{section44}
We conduct ablation experiments to validate the necessity of each part in our proposed framework.

\begin{figure}[!t]
   \centering
      \includegraphics[width=0.5\textwidth]{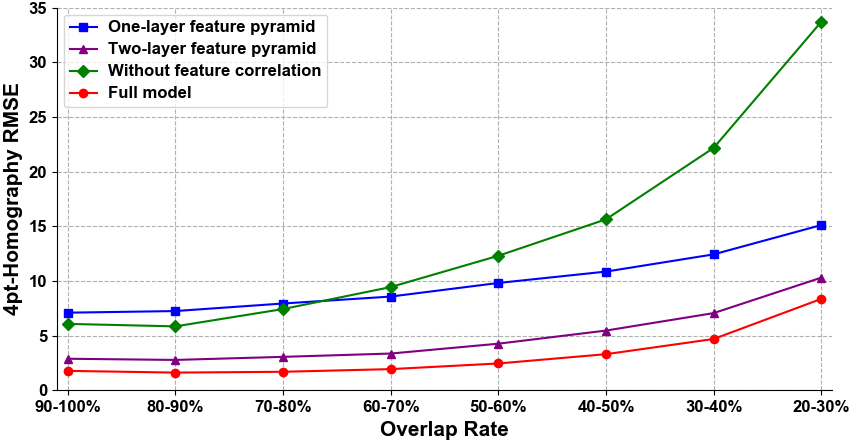}
      \caption{Ablation experiments on feature pyramid and feature correlation for homography estimation.
      Feature pyramid: The three-layer pyramid model is better than one-layer and two-layer. Feature correlation: The model with feature correlation is better than that without.}
      \label{homography-ablation}
\end{figure}

\begin{spacing}{1.5}
\end{spacing}

\noindent\textbf{Feature Pyramid.}
The feature pyramid serves as a multi-scale feature extractor in our method. To reduce parameters, we set the kernel size of each convolutional layer to $3\times 3$. However, the receptive field of the $3\times 3$ kernel is significantly limited. To mitigate this contradiction, the feature pyramid is adopted to extract multi-scale features on different pyramid levels with a fixed kernel size. We evaluate the significance of the feature pyramid with our synthetic dataset on the homography estimation task. As we can see in Fig. \ref{homography-ablation}, our complete pyramid model has significantly smaller errors than one-layer or two-layer models.


\begin{spacing}{1.5}
\end{spacing}

\noindent\textbf{Feature Correlation.}
The feature correlation layer plays the role of feature matching in our method. Different from other deep homography estimations \cite{detone2016deep, nguyen2018unsupervised, zhang2019content, le2020deep} that match features by learning convolutional filters, our feature correlation layers match features by making full use of the features extracted by the convolutional layers. Besides that, our global-to-local strategy ensures our capability to match features all over feature maps. To validate the effects of feature correlation, we experiment with removing feature correlation layers, where both the global correlation and the local correlation are ablated. The results are shown in Fig. \ref{homography-ablation}, where the RMSE increases with a large margin in the absence of feature correlation, especially with the low overlap rate.

\begin{spacing}{1.5}
\end{spacing}

\begin{figure}[!t]
   \centering
   \subfigure[w/o Edge Deformation Branch]
   {\includegraphics[width=0.24\textwidth]{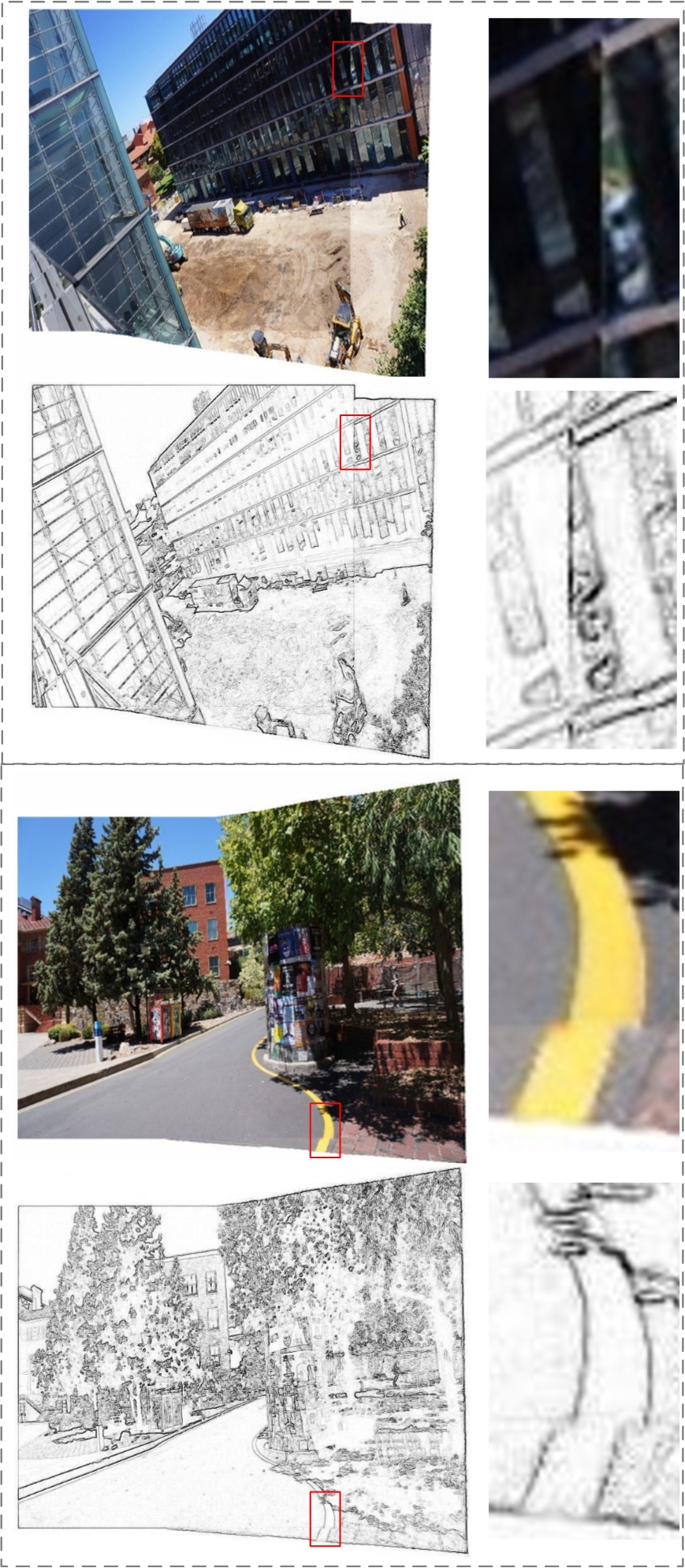}}
   \subfigure[w/ Edge Deformation Branch]
   {\includegraphics[width=0.24\textwidth]{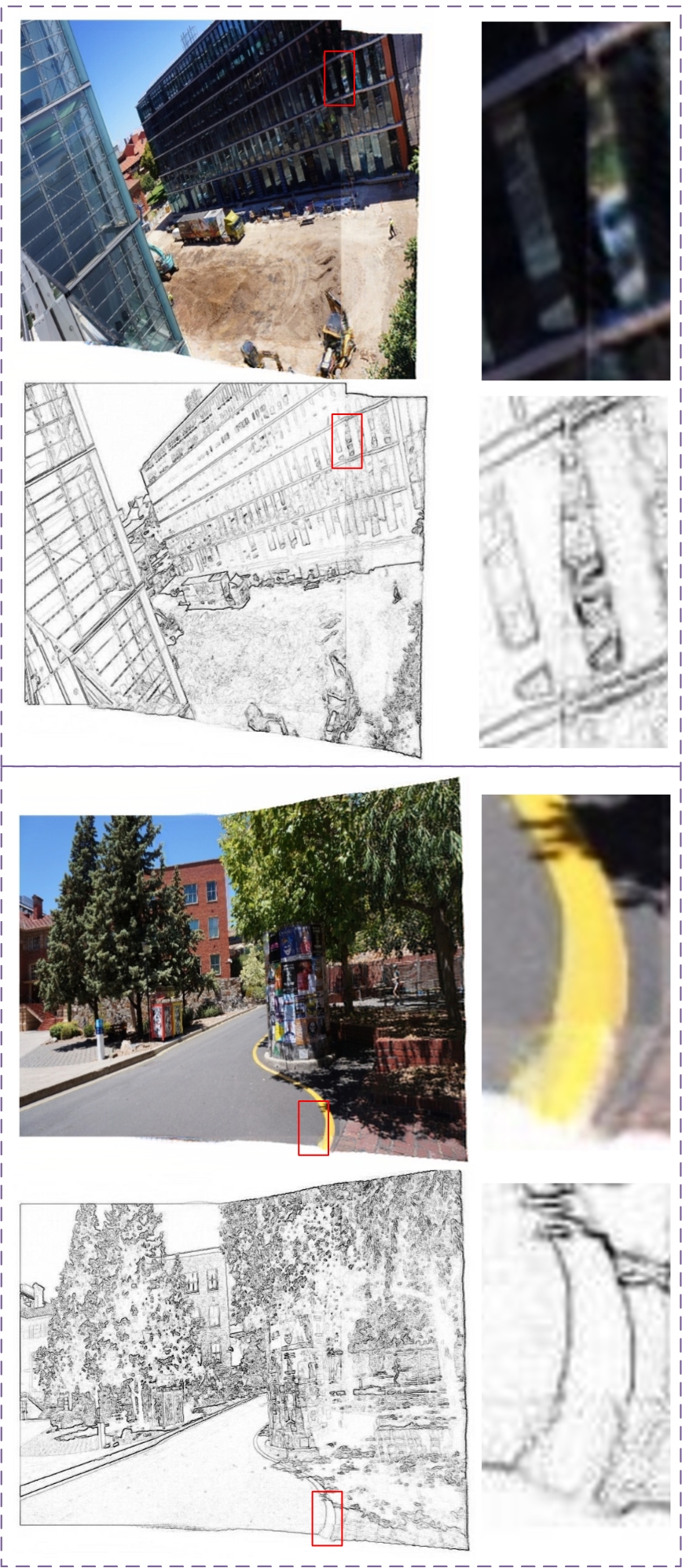}}
   \caption{Ablation experiment on real images to validate the effects of edge deformation branch.}
   \label{ESB_2}
\end{figure}

\noindent\textbf{Edge Deformation Branch.}
In order to validate the effectiveness of the edge deformation branch, we carry out the ablation experiments on real images. We retrain the deformation module without the edge deformation branch. The results are illustrated in Fig. \ref{ESB_2}, and we can observe:

(1) With or without the edge deformation branch, the network can learn to eliminate artifacts in the overlapping area.

(2) After ablating this branch, the edges of the stitched images is not discontinuous as shown in Fig. \ref{ESB_2} (a). With this branch (Fig. \ref{ESB_2} (b)), the network further learns to smooth the discontinuous edges, contributing to visually pleasing and edge-continuity stitched results.

\section{Conclusion}
\label{section5}
This paper presents a novel deep image stitching algorithm that can stitch images from arbitrary views into a perceptually natural image. First, a large-baseline deep homography network is proposed to implement homography estimation and image registration, which outperforms existing deep solutions and traditional solutions with a large margin. Then we present an edge-preserved deformation module to learn the deformation rules of image stitching from the warped images. Furthermore, some schemes are adopted to enable our network the capability of free-size stitching if the full connected layer is inevitable. Experiments show that our method is superior to the existing learning method and shows competitive stitching performance with state-of-the-art traditional methods. Furthermore, as a learning method that is only trained in a synthetic dataset, our method exhibits excellent generalization capability, easily extended to work in other real images.


\normalem
\bibliographystyle{ieeetr}
\bibliography{reference}
\end{document}